\documentclass{article}

\usepackage{amssymb}
\usepackage{latexsym}
\usepackage{natbib}
\usepackage{url}
\usepackage{amsmath}
\usepackage{xcolor}
\usepackage{graphicx,psfrag,epsf}

\begin{document}

\title{Classifying textual data: shallow, deep and ensemble methods}

  \author{Laura Anderlucci, Lucia Guastadisegni, Cinzia Viroli \\
    Department of Statistical Sciences, University of Bologna, Italy
}


  \maketitle

\bigskip

\begin{abstract}
This paper focuses on a comparative evaluation of the most common and modern methods for text classification, including the recent deep learning strategies and ensemble methods. The study is motivated by a challenging real data problem, characterized by high-dimensional and extremely sparse data, deriving from incoming calls to the customer care of an Italian phone company. We will show that deep learning outperforms many classical (shallow) strategies but the combination of shallow and deep learning methods in a unique ensemble classifier may improve the robustness and the accuracy of ``single'' classification methods.
\end{abstract}

\noindent%

\vfill

\section{Introduction}
Nowadays the increasing and rapid progress of technology and the availability of electronic documents from a variety of sources have made a huge amount of textual data available. Hence, one of the prominent research topics of statistical and machine learning communities is to provide suitable and feasible methods to extract high-quality information from unstructured textual data \citep{lata2018text} for the different purposes of clustering, classification and document retrieval \citep{Khan10areview}.

This work originates from an empirical problem of classification of the content of calls made to the customer service of an important mobile phone company in Italy. The received calls are written down by an operator and classified into relevant classes (e.g. claims or request of information for specific services, deals or promotions). The aim is to provide a strategy to automatically assign new tickets into the pre-defined classes, so as to speed up and to improve the assistance service.

In this work we compare and discuss the most common and modern classification methods for textual data. We conduct our experiments on the ticket data previously described that have the peculiarity to be very short and sparse (after a pre-processing step, the tickets have indeed an average length of only 5 words and, thus, the document-term matrix contains zero almost everywhere).

In the scientific literature there are many studies aimed at reviewing and comparing the different strategies for text classification (see, among the others, \cite{korde2012text,PatraSingh,Mironczuk}). Most of these reviews focus on classical methods, called `shallow' strategies in contrast to the recent deep learning strategies. Deep learning methods are receiving an exponentially increasing interest in the last years as powerful tools to learn complex representations of data. They are basically multi-layer architectures composed by nonlinear transformations able to predict data with high precision \citep{lecun}. In particular, deep neural networks (i.e. deep Feed Forward, Recurrent, Auto-encoder, Convolution neural networks) have demonstrated to be particularly successful in supervised classification problems arising in several fields including textual data analysis \citep{schmidhuber}.

In this work a comprehensive study comparing deep learning strategies with classical classification methods is conducted. We also evaluate an ensemble strategy based on voting of base methods \citep{Mironczuk,ONAN2016232}.
To the best of our knowledge, this is the first empirical analysis which evaluates the effectiveness of ensemble learning algorithms based on shallow and deep learning methods on very short textual data.


The paper is organized as follows. Section 2 presents an overview of the conventional classification methods, the recent deep learning strategies and the ensemble classifiers. In Section 3 the real data problem motivating this work is illustrated. Section 4 describes the results of the analysis. We conclude this work with some final remarks in Section 5.

\section{Classification methods}

In this section a description of supervised classification methods, from the classical classification methods to the more recent deep learning and ensemble strategies is given.

\subsection{Classical (shallow) methods}

\paragraph{Na\"{\i}ve-Bayes}
The Na\"{\i}ve-Bayes (NB) classifier is a well-known probabilistic method \citep{JL95,HY01}. The model assumes the conditional independence among variables. More precisely, given $k$ classes and a document-term matrix of dimension $n \times p$, the probability that a generic document $d$ ($d=1,\ldots,n$) belongs to the class $i$ ($i=1,\ldots,k$) is calculated as:
$$ P(i|d)\propto P(i)\prod_{j=1}^pP(t_{j}|i)$$
where $P(t_{j}|i)$ is the probability that the word $t_{j}$ ($j=1,\ldots,p$) belongs to a document of class $i$, and $P(i)$ is the prior probability that a text document belongs to the class $i$.

The document $d$ is finally assigned to the class $i$ that has the maximum posterior probability \citep{schutze2008introduction}. Depending on the probabilistic choices for $P(i)$ (uniform or weighted) and $P(t_{j}|i)$ (i.e. Bernoulli, Multinomial etc.) several models can be defined.

Thanks to its simplicity, the Na\"{\i}ve-Bayes classifier is computationally straightforward and it can be successfully trained with a small amount of data and high-dimensional complex data \citep{Kumbhar2016ASO}.
Even if in practice the independence assumption may be unrealistic, the Na\"{\i}ve-Bayes classifier performs very well in classifying textual data, where terms are not usually highly correlated.

\paragraph{$k$-Nearest Neighbor}
The $k$-Nearest Neighbor ($k$-NN) is one of the simplest non-parametric method used in classification \citep{Kumbhar2016ASO,CH67}.
In $k$-NN with $k=1$ the distance between a new document and the set of documents of the training set is computed, then the new document is assigned to the class of the nearest document.
In the general $k$-NN case, the document is assigned to the modal class among its $k$ nearest neighbors.
The optimal value of $k$ and the best distance to adopt (e.g. Euclidean, Manhattan, Canberra etc.) can be selected by cross-validation. For the analysis of textual data a prominent measure of distance is based on the cosine similarity, as it considers the non-zero elements of the vectors only, allowing to measure the dissimilarity between two documents in terms of their subject matter. Given two $p$-dimensional documents, say $\textbf{x}$ and $\textbf{y}$, the cosine distance of the two corresponding frequency vectors is:

\begin{eqnarray}\label{eqn:cosdist}
d(\textbf{x},\textbf{y})=1-\frac{\sum_{j=1}^px_jy_j}{\sqrt{\sum_{j=1}^px_j^2}\sqrt{\sum_{j=1}^py_j^2}},
\end{eqnarray}

where $x_j$ and $y_j$ denote the frequency of word $j$ in document $\textbf{x}$ and $\textbf{y}$, respectively. This measure is not affected by the amount of zeros and is a normalized synthesis of the $p$-variate terms of the documents. Since the elements of $\textbf{x}$ and $\textbf{y}$ are positive or null frequencies, it is easy to prove that the distance ranges between 0 and 1.

One of the main disadvantage of $k$-Nearest Neighbor is its computational cost that can be very intensive when the number of features/terms is high, or in case of very large training set.

\paragraph{Component-wise classifier}
The component-wise classifiers consider the distance of each document from a reference value of the classes, separately for each variable. Then the variable-wise distances are simply aggregated across the variables and a new document is assigned to the class from which it has the minimum overall dissimilarity. Within this family of methods there are the Centroid Classifier (CC), the Nearest Shrunken Centroid classifier (NSC), the Median Classifier (MC) and the Quantile Classifier (QC).

Centroid-based classifier \citep{Tibshirani6567} considers the similarity of a text document to the centroid of each class, where the similarity measure is calculated as the ones' complement of the Euclidean distance and the centroid is the vector of the average frequencies of each term across all documents of a specific class.
The new document is assigned to the class with the most similar centroid.
When applied to text classification with TF-IDF transformed frequencies, the centroid classifier is known as the Rocchio classifier \citep{harish2010representation}.

The Nearest Shrunken Centroid Classifier \citep{tibshirani2003class} represents an extension of the centroid classifier, which corrects the centroid of each class by `pushing' it to the general centroid of all the text documents.
The NSC classifier coincides with the Centroid classifier when the `shrinkage' parameter (threshold) is equal to 0.

The Median Classifier is a recent method proposed in order to extend the centroid classifier to problems with heavy tails \citep{Jor04,GHO05}. \cite{HTX09} introduced a component-wise median-based classifier which behaves well in high-dimensional data. A new observed vector is assigned to the class having the smallest $L_1$-distance from the class conditional component-wise median vectors of the training set.
Instead of considering the distance from the `core' of a distribution (mean or median) as the major source of the discriminatory information, \cite{hennig2016quantile} proposed to use generic quantiles depending on percentiles in [0,1]. The so-called Quantile Classifier is thus defined. The value of the percentile can be chosen by cross-validation. The method proved to outperform the centroid and the median classifier in high-dimensional and skew data. When the percentile is 0.5 the Quantile classifier coincides with the Median classifier.

\paragraph{Linear Discriminant Analysis}
The Linear Discriminant Analysis (LDA) is one of the most popular and classical statistical method for classification. Proposed by R.A. Fisher in 1936 \citep{fisher}, the idea is to find the linear combination of the variables that better separates the $k$ classes. In linear discriminant analysis the text documents are projected onto a lower dimensional space, where the classes are well separated and only features that carry information useful for classification are extracted \citep{torkkola2001linear}. LDA can suffer from sparsity in the data: the presence of many zeros may compromise the classification performance as the method is based on linear combinations.

\paragraph{Support Vector Machines}
Support Vector Machines (SVM) are supervised learning tools developed in the context of machine learning based on kernel methods \citep{Cortes1995,WZZ08}. The SVM algorithm finds the linear or nonlinear hyper plane that separates two classes (positive and negative training set) \citep{harish2010representation}.
The optimal hyper plane is the one that gives the maximal margin between the documents of the two classes. The support vector includes the documents that are closest to the decision surface. Different kernel functions can be specified: from the simple linear to nonlinear alternatives, such as polynomial or sigmoid functions \citep{Kumbhar2016ASO}.
Support Vector Machine is accurate also in presence of a high number of features, but can be very time-consuming in training and classifying the documents \citep{Khan10areview}.

\paragraph{Decision Trees and Random Forests}
Decision trees are nonparametric approaches that allow to represent a certain number of classification rules referred to single variables taken one at a time with a tree structure \citep{BFOS84}.
A tree structure is composed by different elements: internal nodes represent the features, each ramification of the structure represents the best split obtainable for a given feature taken as reference (according to a certain criterion) and leaves represent the classes.
The number of branches grows exponentially with the number of features considered. For this reason such method can suffer from an excessive number of features. The main advantage of decision tree is its ease of interpretability, even for non-professionals.
Random forests are an ensemble method which is based on the idea of constructing a multitude of classification trees at the training phase: the predicted class is the mode of the classes of the individual trees \citep{Breiman2001}.
Being an extension of the decision trees, they can be computationally very high-demanding.

\subsection{Deep learning methods}
\begin{figure}
  \centering
  \includegraphics[width=0.9\textwidth]{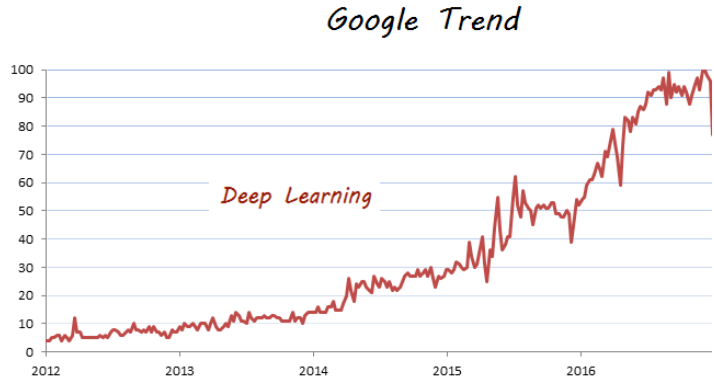}
  \caption{Number of (relative) times the word `Deep Learning'  was searched by Google along the years}\label{fig.cit}
\end{figure}

Deep learning is a prominent research topic in machine learning and pattern recognition. The exponential increasing popularity of the deep learning methods (see Figure \ref{fig.cit}) derives from the fact that they have demonstrated to be powerful and flexible tools to learn complex representations of data in several fields of research. Deep structures are a multi-layer stack of algorithms or modules able to gradually learn a huge number of parameters in an architecture composed by multiple nonlinear transformations \citep{lecun}. Typically, and for historical reasons, a structure for deep learning is identified with advanced neural networks \citep{schmidhuber}.

\paragraph{Neural networks}

Artificial neural networks (ANNs) were developed to mimic the human brain in arriving at a decision. The classical structure of an ANN is an architecture of nonlinearly connected processors called neurons \citep{Lai2015RCN2886521.2886636}.
Neural networks are usually trained with back-propagation, where the parameters of the networks change until an accuracy measure is maximized. The simplest neural networks are composed by only two layers: an input layer representing the observed features and an output layer that gives the predicted classes or values. By adding layers to this hierarchy Deep Neural Networks (DNN) are defined, and they include many relevant models: deep Feed Forward, Recurrent, Auto-encoder, Convolution neural networks are some effective and very used algorithms.
Deep Neural Networks perform very well for classification, even in presence of high-dimensional and noisy features \citep{Khan10areview}.

\subsection{Ensemble methods}

Ensemble strategies are procedures that combine multiple classifiers by considering them as a `committee' of decision makers.
Each classifier is a voting member and the committee produces a final prediction based on
their votes, that could be weighted or unweighed. Ensemble methods can adopt different strategies. Bagging, boosting, AdaBoost, stacked generalisation, mixtures of experts, and voting based methods are some relevant approaches \citep{Mironczuk}.
Ensembles are commonly applied to minimize specific drawbacks, such as overfitting and the curse of dimensionality, and they usually achieve better results than those from single
classifiers \citep{WANG201477,Dietterich}.
Ensemble strategies have been successfully applied in textual data analysis. Recently, \cite{ONAN2016232} presented a comprehensive study of comparing base learning algorithms with some widely utilized ensemble methods; \cite{LOCHTER2016243} proposed an ensemble strategy to combine the state-of-the-art text processing approaches, as text normalization and semantic indexing techniques, with traditional classification methods to automatically detect opinion in short text messages.

\section{Ticket data}
The dataset contains the text of $n=2129$ received calls (tickets). The tickets have then been classified by independent operators to $k=5$ main classes described in Table \ref{tab1}.

\begin{table}[ht]
\caption{Number of tickets for each class.\label{tab1}}
\centering
\begin{tabular}{clc}
  \hline
 Class & Description &Frequencies \\
  \hline
ACT&  Activation of SIM, ADSL, new contracts                        &  407 \\
BAL&  General information about current balance, consumption, etc.  &  471 \\
OFF&  Request of information about new offers and promotions        &  376 \\
TOP&  Top-up                                                        &  435 \\
ISS&  Issues with password, top-up, internet connection, etc.       &  440 \\
   \hline
\end{tabular}
\end{table}

Raw data were processed via lemmatization (it removes inflectional endings of a word and returns its canonic form, which is known as the lemma) and stemming (it reduces inflected or derived words to their unique root, thus removing its derivational affixes) \citep{schutze2008introduction}. Then, some terms have been filtered out in order to remove very common non-informative words and articles in the Italian language (\emph{stopwords}). The stopwords can be viewed as noise since they do not vary significantly among classes \citep{PatraSingh}. Moreover, the terms that appear only once in the whole observed dataset have been removed since considered too rare to be significantly useful for classification.

After pre-processing, data are represented as Bag-of-words \citep{harish2010representation} in a final document-term matrix of dimension $n \times p$ with $p=489$ and $n=2129$. Each cell contains the frequency of a term in a text document. The peculiarity and major challenge of this dataset is the limited number of words used, on average, for each ticket. In fact, after pre-processing, the documents have an average length of 5 words. As previously observed, this determines a sparse matrix with many zeros, and this could affect the classification performance of many methods.

In a cross-validation scheme we evaluated alternative weighting schemes \citep{schutze2008introduction} to the observed frequencies such as the absolute frequencies adjusted according to the TF-IDF transformation, or with the average frequency of the terms in the documents that contain them, the relative frequencies, and the relative TF-IDF transformation. For most classification methods the alternative weighting schemes do not offer an advantage on the classification with respect to the observed frequencies.

\section{Experimental results on Ticket document collection}

The classification methods presented in Section 2 have been implemented in a 50-fold cross-validation study \citep{kohavi1995study}. At each step, each algorithm is evaluated in terms of the accuracy rate, i.e. the number of correctly classified documents over the total number of documents.

\paragraph{Na\"{\i}ve-Bayes}

The method has been implemented with the R package \texttt{quanteda}. Different settings have been considered:
multinomial or Bernoulli conditional distribution, uniform prior distribution or priors proportional to the class sizes,
model with or without smoothing parameter ($\lambda$ equal to 1 or 0). Table \ref{tab2} contains the accuracy for the different settings. Classification performances are generally good with the exception of the Bernoulli probabilistic model without smoothing. The best model model is the Bernoulli Na\"{\i}ve-Bayes with priors proportional to the class sizes and $\lambda=1$.

\begin{table}[ht]
\caption{Accuracy rates (multiplied by 100) of the Na\"{\i}ve-Bayes classifier. In brackets cross-validation standard errors (multiplied by 100) are reported.\label{tab2}}
	\begin{center}
		\begin{tabular}{lllc}
	\hline
	\textbf{Probabilistic} & \textbf{Prior} & \textbf{Smoothing} & \textbf{Accuracy} \\
\textbf{model} &  & \textbf{Parameter} &  \\
	\hline
Multinomial & uniform  & $\lambda=1$ & 98.27 \ (0.28)  \\
Multinomial & class document frequency & $\lambda=1$ & 98.07 \ (0.31)\\
Multinomial & uniform  & $\lambda=0$  & 89.76 \ (0.61) \\
Multinomial & class document frequency & $\lambda=0$ & 89.67 \ (0.63)\\
Bernoulli & uniform  & $\lambda=1$  & 98.50 \ (0.28)\\
\textbf{Bernoulli} & \textbf{class document frequency} & $\mathbf{\lambda=1}$  & \textbf{98.54} \ \textbf{(0.28)}\\
Bernoulli & uniform & $\lambda=0$  & 42.68 \ (0.79)\\
Bernoulli & class document frequency & $\lambda=0$  &42.68 \ (0.79)\\
\hline
\end{tabular}
	\end{center}
\end{table}

\paragraph{$k$-Nearest Neighbor}
The $k$-NN has been applied for a number of neighbors $k=1,2,3$, with several distance measures: the Euclidean (L2), Manhattan (L1) and Canberra distances have been computed with the R package \texttt{stats}, the Cosine distance via the R package \texttt{skmeans}.
The accuracy of the method according to the different settings is shown in Table \ref{tab3}. As expected, due to the sparsity of the data, the distance that produces the best classification is the Cosine dissimilarity.

\begin{table}[ht]
\caption{Accuracy rates (multiplied by 100) of the $k$-Nearest Neighbor classifier for different values of $k$ and distances. In brackets cross-validation standard errors (multiplied by 100) are reported.\label{tab3}}
	\begin{center}
		\begin{tabular}{llc}
	\hline
	 $\textbf{k}$ & \textbf{Distance} & \textbf{Accuracy} \\
	\hline
	1-NN & Euclidean distance & 97.33 \ (0.33)\\
	2-NN & Euclidean distance & 94.07 \ (0.47) \\
	3-NN & Euclidean distance &  91.32 \ (0.67) \\
	1-NN & Manhattan distance & 97.28 \ (0.34) \\
	2-NN & Manhattan distance & 94.31 \ (0.45) \\
	3-NN & Manhattan distance & 91.51 \ (0.67) \\
	1-NN & Canberra distance & 98.09 \ (0.28) \\
	2-NN & Canberra distance & 96.30 \ (0.33) \\
	3-NN & Canberra distance & 95.97 \ (0.42) \\
	\textbf{1-NN} & \textbf{Cosine distance }& \textbf{98.42 }\ \textbf{(0.28)} \\
	2-NN & Cosine distance & 96.63 \ (0.36) \\
	3-NN & Cosine distance & 96.26 \ (0.38) \\
\hline
\end{tabular}
	\end{center}
\end{table}

\paragraph{Component-wise classifier}

The Centroid classifier and the Nearest Shrunken Centroid classifier \citep{Tibshirani6567} have been implemented with the R package \texttt{pamr}. NSC has been computed with threshold parameter equal to 0.5, 1 and 2.
The Quantile Classifier has been implemented with the R package \texttt{quantileDA} for 100 different percentiles (in Table \ref{tab4} only the higher accuracy rate corresponding to the percentile 0.98 is reported). The Median Classifier is obtained as special case of the Quantile one for the percentile 0.5. Results are shown in Table \ref{tab4}. Classification results are not as good as the previous methods. The component-wise classifiers suffer from the high sparsity of the data more than other strategies. This is due to the fact that in most cases the centroid, the median and even the extreme quantiles are zero.

\begin{table}[ht]
\caption{Accuracy rates (multiplied by 100) of different component-wise methods. In brackets cross-validation standard errors (multiplied by 100) are reported.\label{tab4}}
	\begin{center}
		\begin{tabular}{lc}
	\hline
	 \textbf{Method} & \textbf{Accuracy} \\
	\hline
	\textbf{CC} & \textbf{95.40} \ \textbf{(0.43)} \\
    NSC (threshold=0.5)& 92.95 \ (0.49) \\
	NSC (threshold=1) & 90.27 \ (0.52) \\
	NSC (threshold=2) & 87.24 \ (0.60) \\
	MC & 48.05 \ (0.59) \\
	QC (percentile 0.98)  & 91.34 \ (0.59)\\
\hline
\end{tabular}
	\end{center}
\end{table}

\paragraph{Linear Discriminant Analysis}
Fisher's LDA has been implemented by the R package \texttt{MASS}. The accuracy obtained by cross-validation is 97.56\% with a standard error (multiplied by 100) of 0.27.

\paragraph{Support Vector Machines}
Support vector machines have been implemented with the R package \texttt{e$1071$} with different types of kernels: linear, radial base, polynomial and sigmoid. As shown in Table \ref{tab5}, the best model is the SVM with sigmoid kernel.

\begin{table}[ht]
\caption{Accuracy rates (multiplied by 100) of Support Vector Machines with different kernels. In brackets cross-validation standard errors (multiplied by 100) are reported.\label{tab5}}
	\begin{center}
		\begin{tabular}{lc}
	\hline
	 \textbf{Kernel} & \textbf{Accuracy} \\
	\hline
	Linear kernel & 98.13 \ (0.24) \\
	Polynomial kernel & 58.85 \ (1.02) \\
	Radial base kernel & 91.40 \ (0.66) \\
	\textbf{Sigmoid kernel} &\textbf{ 98.97} \ \textbf{(0.20)}\\
\hline
\end{tabular}
	\end{center}
\end{table}

\paragraph{Decision Trees and Random Forests}
We implemented the classification trees by the R package \texttt{rpart}, with the best combination of tuning parameters \texttt{minsplit=}(5, 10, 15, 20) and  \texttt{cp=} (0.001, 0.01, 0.1, 0.2) chosen in cross-validation.
Random Forests \citep{Breiman2001} have been computed with \texttt{ntree=}500 and tuning parameter \texttt{mtry} selected with the function \texttt{tuneRF}. We evaluated also Random Forests for high dimensional data proposed by \cite{xu2012classifying}, a special algorithm that can classify very high-dimensional data with random forests built using small
subspaces. They have been implemented with parameter \texttt{mtry=}50.

\begin{table}[h]
\caption{Accuracy rates (multiplied by 100) of Classification Trees and Random Forests. In brackets cross-validation standard errors (multiplied by 100) are reported.\label{tab6}}
	\begin{center}
		\begin{tabular}{llc}
	\hline
	 \textbf{Method} & \textbf{Accuracy} \\
	\hline
	Decision trees & 93.24 \ (0.49) \\
	\textbf{Random Forests} & \textbf{97.85 }\ \textbf{(0.27)} \\
	Random Forests for high dimensional data  & 96.95 \ (0.32)\\
\hline
\end{tabular}
	\end{center}
\end{table}

\paragraph{Neural Networks and Deep Neural Networks}	

Neural networks have been applied to the data with different structure chosen semi-automatically by the software with the R package \texttt{nnet}, with a number of nodes equal to 1 or 2.
Deep neural networks \citep{Lai2015RCN2886521.2886636} have been implemented under an R interface to Python called \texttt{keras}. We considered several architectures: 1 or 2 hidden layers, combinations of different activation nonlinear functions at the different layers (relu, softmax, etc.), different number of nodes and batch sizes. In the following table, we present results for the best architectures, within the many combinations of tuning parameters, in the family of classical neural networks and deep neural networks with 1 and 2 hidden layers.
Results indicate how a deep structure can largely improve the classification. One hidden layer in this case is enough compared to the case of 2 hidden layers or more. The best DNN model with 1 hidden layer corresponds to the following setting for the tuning parameters: unit=10, activation=`selu', optimizer=`adam', Epoch=100, batch size=32, validation split=0.2.

\begin{table}[ht]
\caption{Accuracy rates (multiplied by 100) of ANNs and Deep NNs. In brackets cross-validation standard errors (multiplied by 100) are reported.\label{tab6}}
	\begin{center}
		\begin{tabular}{lc}
	\hline
	\textbf{Method} & \textbf{Accuracy} \\
	\hline
	Neural Network (1 node) & \ 70.32 \ (1.63) \\
	Neural Network (2 nodes) & 90.73 \ (0.86) \\
    \textbf{Deep Neural Network (1 hidden layer)} & \textbf{98.82} \ \textbf{(0.25)} \\
	Deep Neural Network (2 hidden layers) & 98.73 \ (0.24) \\
	\hline
\end{tabular}
	\end{center}
\end{table}

\paragraph{Ensemble of classifiers}	
From the previous results it emerges that different classifiers produce overall good classification. Here we consider the best classifiers with an accuracy of about $98\%$, taking the best setting within each family of models, namely:

\begin{enumerate}
  \item Na\"{\i}ve-Bayes classifier with Bernoulli distribution and priors proportional to the class sizes (accuracy $98.54\%$)
  \item 1-NN with Cosine distance (accuracy $98.42\%$)
  \item Support Vector Machines with sigmoid kernel (accuracy $98.97 \%$)
  \item Random Forests (accuracy  $97.85\%$)
  \item Deep Neural Networks with 1 hidden layer (accuracy $98.82 \%$)
\end{enumerate}

\begin{figure}
  \centering
  \includegraphics[width=\textwidth]{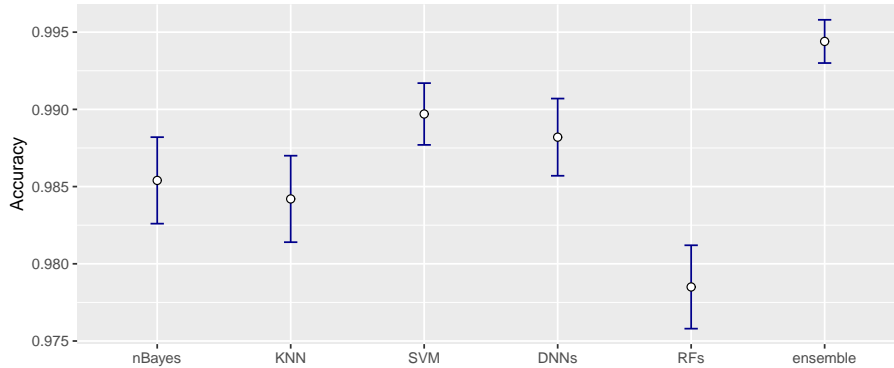}
  \caption{Accuracy obtained by the different strategies in 50-fold cross-validation}\label{fig1}
\end{figure}

The five algorithms are considered as base learners in an ensemble strategy where the aim is to combine the different decisions about the ticket allocation. Here we adopt the majority vote strategy: a new ticket is assigned to the modal class of the five classification strategies. Figure \ref{fig1} depicts the plot of the average accuracy (with confidence intervals) obtained by cross-validation by the 5 best methods and the ensemble strategy. The ensemble classifier has an accuracy of $99.44\%$ with standard error 0.14 (multiplied by 100).

\begin{table}[h]
\caption{Precision, recall and $F$ index for the 5 classes of the base learners and the ensemble classifier}\label{taba}
\centering
\begin{tabular}{lrrrrrr}
 \hline
 \multicolumn{7}{c}{Precision}\\
 & nBayes & KNN & SVM & DNNs & RFs & ensemble \\
  \hline
  ACT & 0.99 & 0.99 & 0.99 & 0.99 & 0.99 & 1.00 \\
  BAL & 0.96 & 0.97 & 0.99 & 0.98 & 0.96 & 0.99 \\
  OFF & 0.97 & 0.95 & 0.97 & 0.99 & 0.96 & 0.99 \\
  TOP & 1.00 & 0.99 & 0.99 & 0.99 & 0.99 & 1.00 \\
  ISS & 1.00 & 0.99 & 0.99 & 0.98 & 0.99 & 1.00 \\
   \hline
 \multicolumn{7}{c}{Recall}\\
 & nBayes & KNN & SVM & DNNs & RFs & ensemble \\
  \hline
  ACT & 0.98 & 0.97 & 0.99 & 0.98 & 0.97 & 1.00 \\
  BAL & 0.98 & 0.98 & 0.97 & 0.99 & 0.99 & 0.99 \\
  OFF & 0.99 & 0.98 & 0.98 & 0.98 & 0.94 & 0.99 \\
  TOP & 0.99 & 0.98 & 1.00 & 0.99 & 0.99 & 1.00 \\
  ISS & 0.98 & 0.99 & 1.00 & 1.00 & 0.99 & 1.00 \\
   \hline
   \multicolumn{7}{c}{F-score}\\
    & nBayes & KNN & SVM & DNNs & RFs & ensemble \\
  \hline
  ACT & 0.98 & 0.97 & 0.99 & 0.98 & 0.97 & 1.00 \\
  BAL & 0.98 & 0.98 & 0.97 & 0.99 & 0.99 & 0.99 \\
  OFF & 0.99 & 0.98 & 0.98 & 0.98 & 0.94 & 0.99 \\
  TOP & 0.99 & 0.98 & 1.00 & 0.99 & 0.99 & 1.00 \\
  ISS & 0.98 & 0.99 & 1.00 & 1.00 & 0.99 & 1.00 \\
  \hline
\end{tabular}
\end{table}

In order to comparatively evaluate the classification accuracy within each class we computed \emph{precision}, \emph{recall} and the $F$-index.
The precision is also called \emph{positive predictive value} because it measures the fraction of correct assignments in a class over the predicted documents of the class. Recall is a measure of sensitivity, since it represents the fraction of correct assignments in a class over the total true documents of the class. Values of precision and recall for the different methods and classes are reported in Table \ref{taba}. The table also contains the so-called $F$-score, a measure that combines precision and recall by their harmonic mean:

$$F=2\cdot \frac{precision \cdot recall}{precision +recall}$$


Results indicate how the ensemble strategy represents a very robust and accurate strategy for this classification task, in terms of the overall accuracy and of the validation measures computed separately for each class.

One may also consider to select classifiers according to the \emph{diversity} between them, as it is recognized to be one of the desired characteristics to improve the performance of an ensemble \citep{ZOUARI20052195}. However, in the classification context, there is no complete and agreed upon theory to explain why and how diversity between individual models contribute toward overall ensemble accuracy \citep{rokach2010ensemble}. Indeed, \cite{kuncheva2003measures} studied the connection between accuracy and ten statistics which can measure diversity among binary classifier outputs. Although there are proven connections between diversity and  accuracy in some special cases, their results raise some doubts about the usefulness of diversity measures in building classifier ensembles in real-life pattern recognition problem.
In addition to these considerations, as in our case there is no evidence of notable diversity (the smallest Yule's $Q$ statistic value is 0.84 for the most accurate classifiers), there is not point in pursuing such direction.

\section{Conclusions}
This paper gives a literature review of the most common classification methods for text classification. Classical classification methods, modern deep learning and ensemble strategies have been considered.

This comprehensive study has been motivated by a real textual data problem, that is the implementation of an automatic process to allocate incoming calls to their correct macro-topic in order to provide a better and fast assistance service. Data have a very complex and high-dimensional structure, caused by the huge number of tickets and terms used by people that call the company for a specific request and by a relevant degree of sparsity. However a list of 5 classifiers have been identified as powerful tools for the data classification. Among these, the prominent deep neural networks, which many researchers consider to be the only true magic recipe for prediction and classification. The different methods have pros and cons. For instance Naive-Bayes has the great advantage to be computationally straightforward compared to the others methods and to have a high accuracy classification rate. Deep Learning strategies are very flexible and work very well in all most situations at the price of demanding computational burden.

The best classification performances are obtained by Support Vector Machines with sigmoid kernel, Naive-Bayes classifier with Bernoulli distribution, the 1-Nearest Neighbor classifier with cosine distance, Random Forests and Deep Neural Networks.

The ensemble strategy we proposed seems to be robust and is more accurate of the classification methods separately considered. In fact, the final ensemble outperforms all the base classifiers, even the deep learning methods, yielding only a small fraction of misclassified tickets. Implementation of such strategy is straightforward but powerful, as it combines the good individual results and the specific methodological features for an overall better performance.

\bibliographystyle{chicago}
\bibliography{biblio}

\end{document}